\documentclass[letterpaper, 10 pt, conference]{ieeeconf}  

\IEEEoverridecommandlockouts                              

\usepackage{graphics} 
\usepackage{epsfig} 
\usepackage{subfigure}
\usepackage{amsmath} 
\usepackage{amssymb}  
\usepackage{amsfonts}

\usepackage{amsthm}
\usepackage{gensymb}
\usepackage[utf8]{inputenc}
\usepackage[english]{babel}
\usepackage{color}
\usepackage{mathtools}
\usepackage{algorithm}
\usepackage[noend]{algpseudocode}
\usepackage[font=footnotesize]{caption}
\usepackage{tabularx}
\usepackage{bm}
\usepackage{multirow}
\usepackage{cite}
\usepackage{xcolor}
 \usepackage{booktabs}

\theoremstyle{plain}

\theoremstyle{definition}

\theoremstyle{remark}

\usepackage[normalem]{ulem}

\title{\LARGE \bf Robotic Lime Picking by Considering Leaves as Permeable Obstacles\vspace{-1 ex}}
\author{Heramb Nemlekar$^*$, Ziang Liu$^*$\thanks{$^*$ Denotes equal contribution. Heramb Nemlekar (nemlekar@usc.edu) is the corresponding author.}, Suraj Kothawade, Sherdil Niyaz, Barath Raghavan and Stefanos Nikolaidis~\\University of Southern California} 

\begin{document}

\bstctlcite{references:BSTcontrol}

\maketitle
\thispagestyle{empty}
\pagestyle{empty}

\begin{abstract}
The problem of robotic lime picking is challenging; lime plants have dense foliage which makes it difficult for a robotic arm to grasp a lime without coming in contact with leaves. Existing approaches either do not consider leaves, or treat them as obstacles and completely avoid them, often resulting in undesirable or infeasible plans. 
We focus on reaching a lime in the presence of dense foliage by considering the leaves of a plant as \textit{permeable obstacles} with a collision cost. 
We then adapt the rapidly exploring random tree star (RRT*) algorithm for the problem of fruit harvesting by incorporating the cost of collision with leaves into the path cost. To reduce the time required for finding low-cost paths to goal, we bias the growth of the tree using an artificial potential field (APF). We compare our proposed method with prior work in a 2-D environment and a 6-DOF robot simulation. Our experiments and a real-world demonstration on a robotic lime picking task demonstrate the applicability of our approach.
\end{abstract}


\section{Introduction}\label{sec:intro}

Fruit harvesting is a labor-intensive task that constitutes a significant portion of the fruit production cost~\cite{harrell1987economic, whitney1995review, jukema2009arbeidskosten}. Introducing robots can be a cost-effective solution for labor shortages in fruit harvesting \cite{sario1993robotics}. The task of fruit picking involves locating the fruit and planning the motion of a robot arm to reach, grab, and retrieve the fruit. To use robots for harvesting fruit, the robot must not damage the plant or the fruit while working as fast as possible~\cite{bontsema1999automatic}. In this work, we focus on the problem of planning the reach-to-grasp motion of a robotic arm for lime picking such that it avoids collision with leaves if possible, but will push through the leaves if there is no alternative.



Previous research in robotic fruit harvesting uses rapidly exploring random tree (RRT) or its variants \cite{cao2019rrt, lehnert2018pepper, yang2017obstacle, hemming2014robot} to plan the reach-to-grasp motion of fruit-picking robots. These robots have high degrees of freedom (DOF) arms to reach and grab the fruit. Hence, sampling-based methods are preferred as they are less computationally expensive than search-based methods like A* in high dimensions \cite{paulin2015grape}. 
While planning, the grasp pose for a target fruit is considered as the goal, and the stems and untargeted fruits are considered as obstacles. 
Previous work also avoids collision with the leaves \cite{van2003collision} or models the plant foliage as a cylinder that must be entered radially \cite{baur2014apf, schuetz2015evaluation}. However, the majority of prior work does not consider the leaves at all \cite{van2004line, nguyen2013task, bac2016analysis, cao2019rrt, xie2019simulation}.

This is reasonable when the fruit is well separated from the plant foliage, but in the case of lime trees and many other important subtropical and tropical tree crops, the fruits are often engulfed by leaves~\cite{morton1987fruits}. Ideally, the robot would measure and avoid collision with leaves, avoiding tree damage, while picking fruit. 
However, this ideal is seldom possible in reality. While some fruits may grow in their outer canopy, most fruits are buried deep within in a thicket of leaves and inner branches.\footnote{There is a fundamental, biological reason for this: in the case of citrus, avocado, and many warm-climate crops, the trees develop dense canopies to prevent bark and fruit damage from intense sunshine.}
Thus, there may not be a completely collision-free path to pick a fruit.

\begin{figure}[t]
\centering
\includegraphics[width=0.4\linewidth]{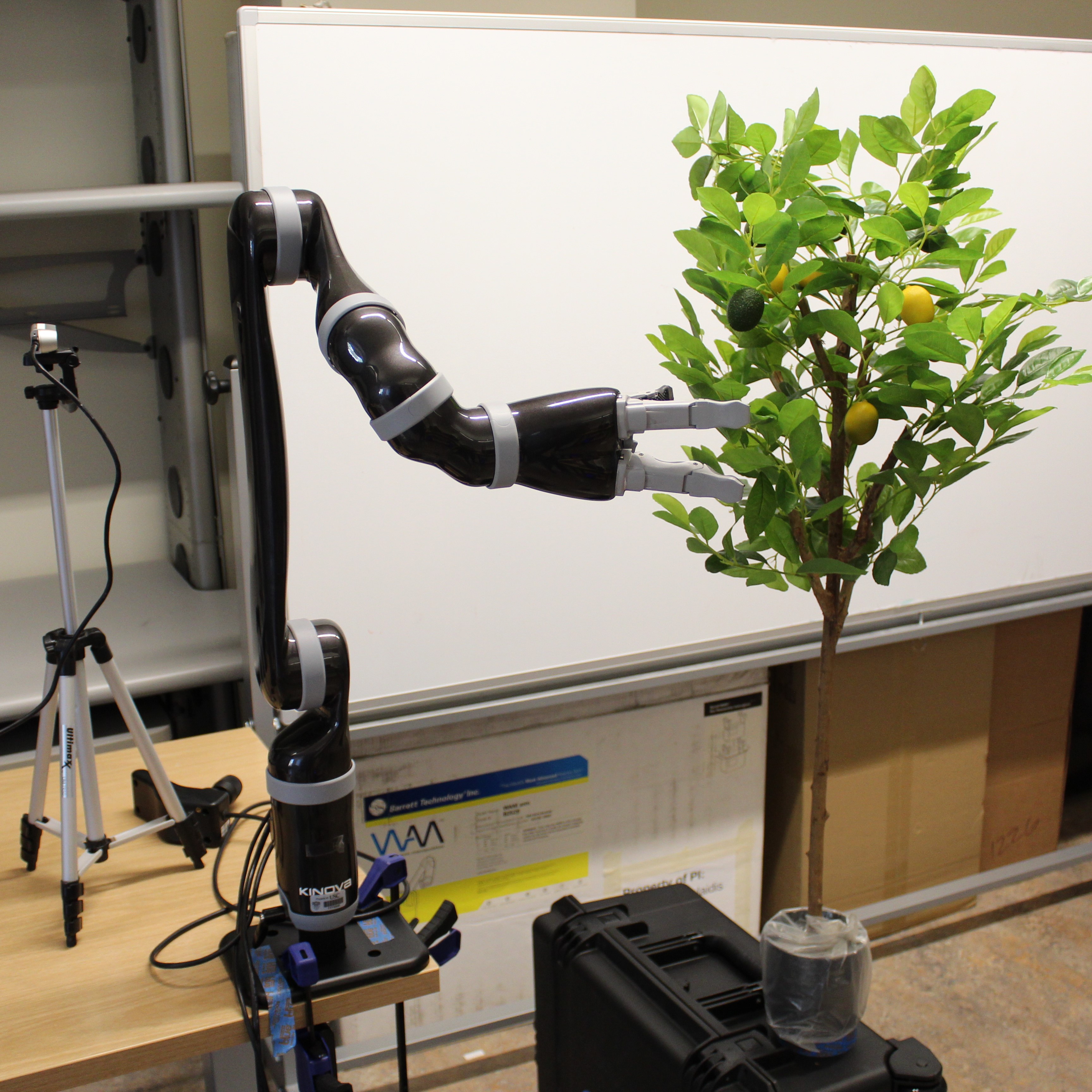}
\quad
\includegraphics[width=0.4\linewidth]{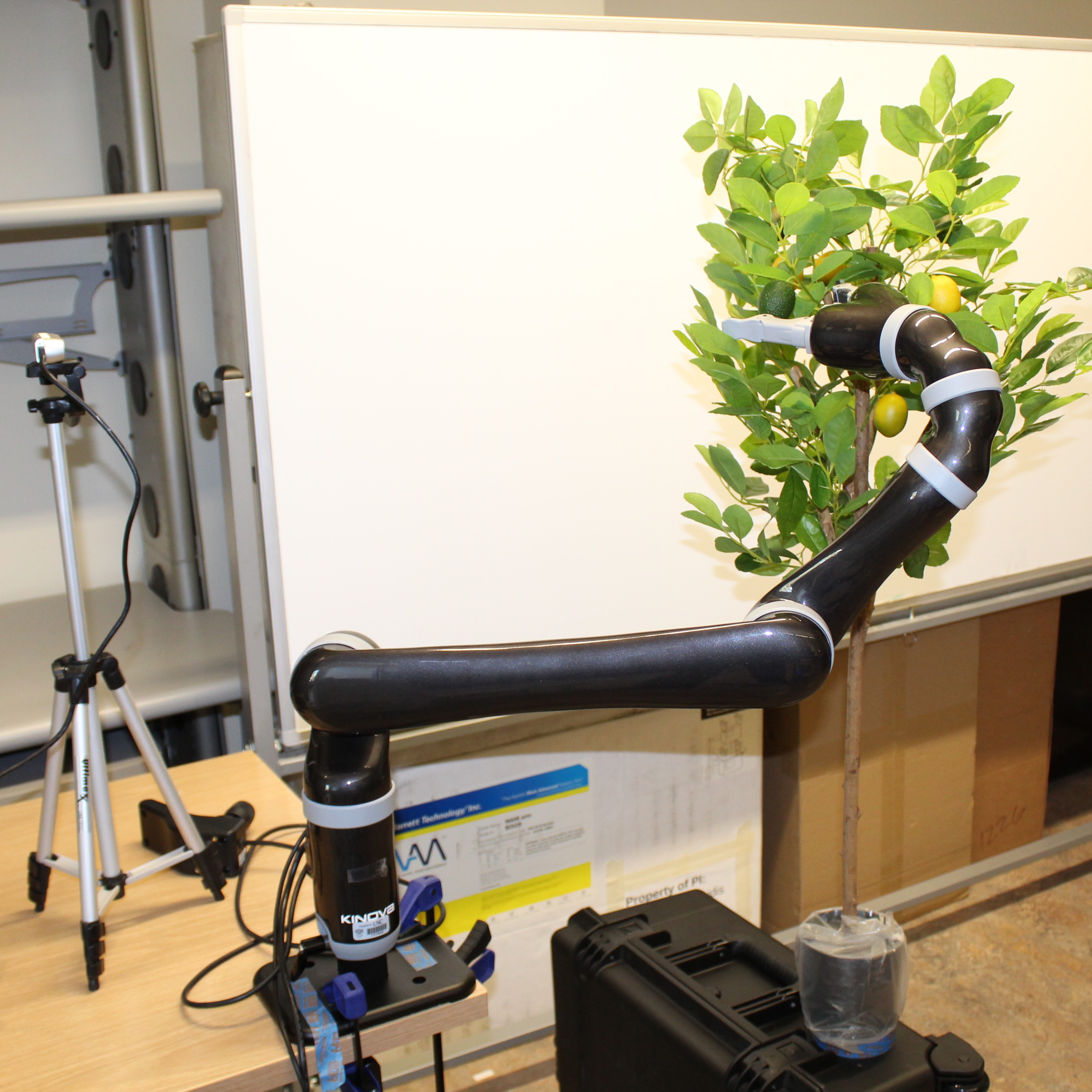}
\caption{Limes are often engulfed by leaves. If a collision-free path does not exist,  the robot must find a path that has minimal collision with the leaves.}
\label{fig:robot_lime_picking}
\vspace{-2 ex}
\end{figure}

Our goal is to find a path that allows for collisions with the leaves but avoids them to the extent possible. To this end, we consider leaves as \textit{permeable obstacles} such that we allow the robot to collide with the leaves but impose a penalty or cost for the collision.
Therefore, we want to find a path to the goal that incurs the least-collision cost. Moreover, if multiple paths have similar collision costs, among these we want to find the shortest one. Thus, we consider the multi-objective problem of finding the shortest, least-collision path.


We can formulate the objective as a weighted sum of the different costs \cite{schuetz2015evaluation}. To optimize this objective,
we adapt the sampling-based algorithm - RRT*, by incorporating the cost of collision with permeable obstacles into the path cost. Therefore, when adding a node into the tree, we choose a parent based on the sum of the path length and collision costs. We also rewire nodes based on this total cost. So as the tree grows, we are able to find paths with lower costs (if available), thus reducing collision with the leaves.

RRT* can require many iterations to find close-to-optimal paths for high-DOF arms, which may be unsuitable for the short cycle time required for fruit harvesting. To speed up the search, one approach is to bias the random samples towards the goal by applying an artificial potential field (APF) and to extend the tree towards the biased sample \cite{qureshi2017prrt*}. However, this may not be ideal for fruit picking where the obstacles (leaves) engulf the goal (lime) and prevent the random sample from being biased towards it.

Instead, we propose applying the potential field bias directly to the nearest node and adaptively changing the bias based on its vicinity to obstacles. Our method extends the node towards the goal when in free space (exploitation) and towards a random sample when near obstacles (exploration). 
Therefore, we can bias paths towards the limes at the start and explore when near the leaves, to effectively find low-cost paths that have minimal collision with leaves. 

In summary, our key insight is considering the leaves as permeable obstacles while planning the reach-to-grasp motion of a robot arm for picking limes. Our work makes the following contributions: (1) We adapt RRT* to account for the cost of collision with leaves during planning. (2) We propose a new artificial potential field (APF) approach for effectively biasing the growth of RRT* to find low-cost paths. We compare our proposed method with prior work in a 2-D environment and a 6-DOF robot simulation. Finally, we show how our method can be practically deployed in a real-world 6-DOF robotic lime picking demonstration. 

\section{Related Work}\label{sec:related}

\subsection{Motion planning for fruit harvesting}

Robot fruit harvesting problems typically require planning the motion of a high-DOF robot arm from its initial configuration to the fruit position through a cluttered environment. We can either plan the robot's path in the workspace \cite{schuetz2015evaluation, baur2014apf} and then map it to the configuration space (C-space) of the robot, or directly plan in the C-space \cite{xie2019simulation, van2002autonomous, van2003collision}.



Sampling-based methods such as RRTs \cite{cao2019rrt, lehnert2018pepper, yang2017obstacle, hemming2014robot}, which use random samples in the C-space to grow the search tree, are typically preferred for high-dimensional planning problems like fruit harvesting. A comparison of motion planners for grape pruning with a 6-DOF robot arm \cite{paulin2015grape} shows that RRT variants have the best overall performance compared to other sampling-based algorithms including KPIECE\cite{sucan2008kpiece} and its variants, EST\cite{Hsu99pathplanning}, and SBL\cite{sanchez2003sbl}. RRTs have also been effectively used for picking litchis \cite{cao2019rrt}. To reduce planning time, the growth of the RRT is biased towards the target litchi using an attractive potential field. Experiments in apple harvesting \cite{nguyen2013task} have shown that RRT-Connect \cite{kuffner2000rrt} is more efficient than KPIECE and EST in finding a path to the target apple. Bi-directional RRTs have also be used for planning the motion of the end-effector for sweet pepper harvesting \cite{bac2016analysis}.


This motivates us to consider RRT-based approaches for the lime picking task. While most approaches for fruit harvesting only consider the main stem, branches or other fruits as obstacles \cite{van2004line, nguyen2013task, bac2016analysis, cao2019rrt, xie2019simulation}, they do not account for collision with the leaves. If the robot ignores the leaves, it could potentially damage the plant foliage.


\subsection{Cost-based RRT approaches}

We can account for collision with the leaves by assigning a cost for the collision. This results in a planning problem where each robot configuration has an associated collision cost and the aim is to find a low-cost path to the goal. 

A common sampling-based approach to cost-based planning is Transition-based RRT (T-RRT) \cite{jaillet2008transition, jaillet2010sampling} and its variants \cite{berenson2011addressing, devaurs2013enhancing, kabutan2018motion}. In T-RRT, if a new node has higher cost than its parent, it is added to the tree with a probability inversely proportional to the difference between the cost of the node and its parent.
However, as this approach rejects most of the nodes with a higher cost, a lot of iterations are required to plan through high-cost regions. 
Moreover, though it can produce low-cost paths, this approach does not reason about searching for paths of decreasing cost.

To find lower-cost paths based on multiple metrics,
e.g. time spent by the planner and path length, the costs can be incorporated into the edge weights of a roadmap graph~\cite{dellin2016completing}. 
In the case of lime picking, we consider the bi-criteria optimization problem that combines \textit{permeable collision cost} and path length. Towards this end, we propose a modified RRT* algorithm~\cite{karaman2011sampling} that uses this combined objective to evaluate the cost of paths in the tree.

\subsection{Potential guided RRT*}

RRT* can require a lot of iterations to search for a least-cost path in the C-space of a cluttered environment. However, we may need to limit the number of iterations so that the lime picking operation can be completed in an economical time \cite{bontsema1999automatic}. Therefore, we want to bias the growth of the tree such that low-cost paths can be found in fewer iterations.
Prior work either samples nodes only from a connectivity region between the start and goal \cite{noreen2018optimal, wang2020improved, mohammed2020rrt}; or uses an APF \cite{qureshi2013potential, qureshi2017prrt*, koukuntla2019deep} to bias the growth of the tree towards the goal. To further improve the efficiency, an APF bias can also be used with a bi-directional RRT* approach\cite{tahir2018potentially, xinyu2019bidirectional, wu2019biased, wang2019prrtconnect}. 

In the majority of prior work, APF is used to bias the random sample towards the goal. Then, the nearest node in the tree is extended towards this biased random sample. However, in the case of lime plants, this would not effectively bias the search towards limes that are engulfed by leaves.

Instead, we can use the APF to directly bias the nearest node of the tree. This requires us to balance between moving in the direction of the random node (exploration) and moving in the direction of the potential force (exploitation) to avoid being stuck in local minima. Prior work implements this approach for an RRT algorithm and uses equal weights for the random and potential force directions \cite{yang2018environmental}. This is not ideal for lime picking, since to effectively find low-cost paths we should be able to explore when near the leaves and exploit when away from the leaves. In this work, we extend this approach to RRT* and propose an adaptive approach for reducing the potential bias when the node is surrounded by obstacles. This allows us to bias the robot towards the limes at the start, and prefer exploration when it is near the leaves to find lower cost paths.


\section{Methodology}\label{sec:method}

Our key insight is to consider the leaves of the plant as \textbf{permeable obstacles}. We define collision with permeable obstacles similar to a soft constraint, where we allow the robot to be in collision with the permeable obstacle but impose a penalty or cost for the collision and try to find a path that minimizes the incurred cost. On the other hand, we consider other obstacles like the plant stem as \textbf{impermeable obstacles} and define collision with impermeable obstacles as a hard constraint that must be avoided. Therefore, we have two types of obstacles: 
$$O = \{O_{permeable}, O_{impermeable}\}$$

We adapt the RRT* algorithm to find a path that avoids impermeable obstacles and minimizes collision with permeable obstacles. Moreover, we apply an artificial potential field (APF) to effectively bias the growth of the tree such that the minimum cost paths can be found in fewer iterations. Our proposed cost-based APF-RRT* method follows similar steps to the vanilla RRT* algorithm. The steps which we modify to incorporate collision cost and apply the potential bias are shown in bold  (see Algorithm \ref{algo:method}).

\begin{algorithm}
\caption{Cost-based APF-RRT* Algorithm}
\begin{algorithmic}[1]
\Require $Q_{free}$, Obstacles $O$, Goal $q_{goal}$
\State add $q_{start}$ to tree $T$
\For{max\_iterations}
    \State $q_{rand}$ $\leftarrow$ random\_sample($Q_{free}$) \label{step:rand}
    \State $q_{near}$ $\leftarrow$ nearest\_node($T$, $q_{rand}$) \label{step:near}
    \State \textbf{$q_{new}$ $\leftarrow$ potential\_biased\_extend($q_{near}$, $q_{rand}$, $\delta$)} \label{step:apf}
    \If{collision\_free($q_{new}$, $O$)}
        \State \textbf{set\_collision\_cost($q_{new}$, $O$)} \label{step:cost}
        \State add $q_{new}$ to $T$
        \State $N$ $\leftarrow$ neighbours($q_{new}$) 
        \State \textbf{update\_parent($q_{new}$, $N$)} \label{step:parent}
        \State \textbf{rewire($q_{new}$, $N$)} \label{step:rewire}
        \State path $\leftarrow$ check\_solution($q_{new}$, $q_{goal}$)
    \EndIf    
\EndFor
\end{algorithmic}
\label{algo:method}
\vspace{-0.5 ex}
\end{algorithm}

\subsection{Incorporating the cost of permeable obstacles} \label{sec:cost}

The goal of the robot planner is to find a path $\tau$ in the collision-free C-space $Q_{free}$ from the start configuration $q_{start}$ of the robot to the goal configuration $q_{goal}$. Since we want to allow the robot to plan through the permeable obstacles, we define $Q_{free} = Q\backslash Q_{impermeable}$.
Therefore, we consider the configurations that lie on the permeable obstacles to belong to $Q_{free}$. 

At each iteration we randomly sample a configuration $q_{rand}$ from the collision-free configuration space $Q_{free}$ (Line \ref{step:rand}). We then find the nearest node $q_{near}$ in the tree $T$ to the random sample based on distance $d$ between the configurations (Line \ref{step:near}). The C-space distance $d$ is the same as in the vanilla RRT*. Here we do not add the collision cost to $d$, since prior work \cite{lee2008cost} has shown that it can cause the planner to only extend nodes near the start (in free space) and not explore the obstacle region (leaves) near the goal.


While the vanilla RRT* algorithm extends the nearest node simply towards the random sample by a step size $\delta$, we extend the nearest node in a \textit{potential-biased direction} by $\delta$ to obtain a new node $q_{new}$ (Line \ref{step:apf}). The potential bias adaptively combines moving in the random direction with moving in the direction of the APF (see Section \ref{sec:apf}). 

Once we obtain $q_{new}$, we then check if the new node is valid i.e. if $q_{new} \in Q_{free}$. This is how the hard constraint that the robot must not be in collision with the impermeable obstacles is imposed. If the node is invalid, we reject the node and sample a new random node (Line \ref{step:rand}). If the node is valid, we check if it lies on a permeable obstacle and assign a collision cost $C_{perm}(q_{new})$ (Line \ref{step:cost}).

\begin{equation}\label{eq:cost}
    C_{perm}(q) = \left\{\begin{array}{ll}
        \mathbb{R}_{> 0} & \quad q \in Q_{permeable}\\
        0 & \quad \text{otherwise}
        \end{array}\right.
\end{equation}

We assign a positive cost if the node lies on a permeable obstacle and $0$ cost if the node does not lie on any obstacle. In order to generate paths that try to minimize this collision cost, we consider the collision cost while: (i) determining the best parent for $q_{new}$ from the neighbouring nodes $N$ (Line \ref{step:parent}) and then (ii) rewiring the neighbours (Line \ref{step:rewire}). We determine the neighbouring nodes $N$ purely based on their distances in the configuration space from the new node $q_{new}$. Specifically, neighbours $N$ are all nodes in the tree $T$ that are within a C-space distance of $r \leq \delta$ from $q_{new}$. 


Traditionally RRT* connects the new node to a parent $q_{par}$ such that the cost of the path from the start node $q_{start}$ to $q_{new}$ i.e. $C_{path}(q_{new}) = d(q_{new}, q_{par}) + C_{path}(q_{par})$, is the least from among the neighbouring nodes $N$. Here, we incorporate the collision cost $C_{perm}$ into the path cost $C_{path}$, so that we can select a parent based on the sum of the length of path from $q_{start}$ and the collision costs incurred on that path. Therefore, the path cost of $q_{new}$ from parent $q_{par}$:
\begin{equation}\label{eq:path_cost}
\begin{split}
    C_{path}(q_{new}, q_{par}) = d(q_{new}, q_{par}) + C_{path}(q_{par})\\+ C_{perm}(q_{par})
\end{split}
\end{equation}

This makes the problem multi-objective. Whether we select a parent with the shortest distance from start or with the least collision cost depends on the magnitude of length $d(\tau)$ and collision $C_{perm}(\tau)$ costs. If $C_{perm} > d$, we will give more importance to minimizing the collision cost and if $C_{perm} < d$ more importance will be given to minimizing $d$. In this work, we select a high collision cost $C_{perm}$ such that we prioritize finding paths with the least collision cost.


We select the best parent $q_{par}^{*}$ for $q_{new}$ based on the minimum $C_{path}(q_{new}, q_{par})$ of all $q_{par} \in N$. The next step is to see if setting the new node as the parent of any of the neighbouring nodes, reduces their path cost. Similar to before, we incorporate the collision cost of the new node $q_{new}$ while determining the cost of path $C_{path}(q, q_{new})$ to a neighbouring node $q$ during the rewiring process. 

Therefore, as the tree grows we rewire neighbouring nodes such as to select parents with the least path cost that includes the cost of collision with permeable obstacles. This allows generating paths by connecting a new node to a parent whose path from start has the least total collision cost. Thus, our proposed method can effectively reason about reducing collision with permeable obstacles like leaves.

After all iterations, we select a node near the goal with the least path cost as the parent of $q_{goal}$ and trace a path back to $q_{start}$.

\subsection{Using APF to bias the extension of nearest node} \label{sec:apf}

We want to bias the extension of the tree, such that we can find lower-cost paths in a limited number of iterations. To do this, we must bias the growth of the tree towards the goal so that we do not waste iterations in exploring the free space. At the same time, we need to explore in areas near the leaves, so that we can find alternative paths that may have lower costs. Prior work \cite{qureshi2017prrt*} uses APF to bias the random node, and thus does not consider the potential forces at the nearest node to adapt how it is extended. 
Instead, we propose using the APF to directly bias the nearest node, such that we can adjust the potential bias depending on the nearby obstacles.

Our APF uses the same quadratic attractive and repulsive potential fields (Equations \ref{eq:attract} \& \ref{eq:repulsive}) as in prior work \cite{qureshi2017prrt*}, with $d_{goal}$ as the distance to goal, $d_{obs}$ as the distance from the obstacle and $d_{obs}^*$ as the maximum effective distance of the repulsive field. $K_{att}$ and $K_{rep}$ are the gains for the attractive and repulsive potentials respectively. We calculate repulsive fields from all obstacles $O$ with a different $K_{rep}$ for permeable and impermeable obstacles. 
The potential at any point in the configuration space is the sum of attractive and repulsive potentials.

\begin{equation}\label{eq:attract}
    U_{att} = K_{att} * d_{goal}^{2}
\end{equation}
\begin{equation}\label{eq:repulsive}
    U_{rep} = 
    \begin{cases} 
      \frac{1}{2}K_{rep} (\frac{1}{d_{obs}} - \frac{1}{d^*_{obs}})^2 & d_{obs} \leq d^*_{obs} \\
      0 & d_{obs} > d^*_{obs} 
   \end{cases}
\end{equation}
\begin{equation}
    U_{tot} = U_{att} + U_{rep}
\end{equation}

\begin{figure}[hbt]
\centering
\includegraphics[width=0.485\linewidth]{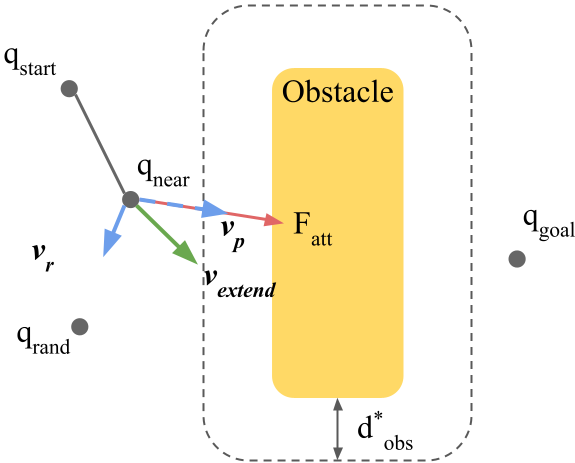}
\includegraphics[width=0.485\linewidth]{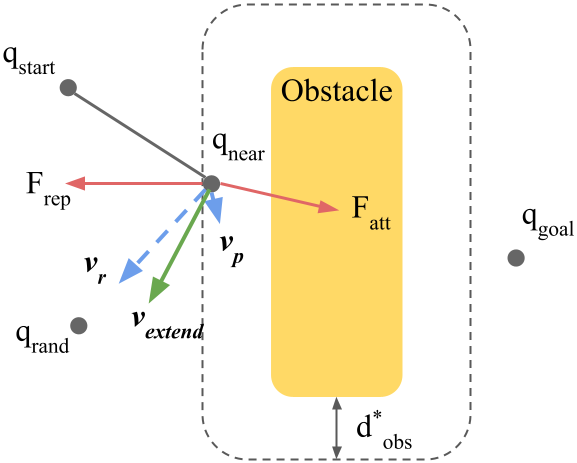}
\caption{Local APF: $q_{near}$ is extended more towards $q_{goal}$ when in free space and towards the $q_{rand}$ when near obstacles. $F_{att}$ and $F_{rep}$ are forces due to $U_{att}$ and $U_{rep}$ respectively. $v_{p}$ is the gradient of $U_{tot}$.}
\label{fig:local_apf}
\vspace{-1 ex}
\end{figure}

The potential bias is applied in Line \ref{step:near} of Algorithm \ref{algo:method} for extending the nearest node $q_{near}$. We calculate the direction ($\bm{v_p}$) of the total potential force at $q_{near}$. $\bm{v_p}$ is equal to the negative gradient of the total potential $U_{tot}$ at $q_{near}$. We also calculate the vector in the direction of the random sample from the nearest node $\bm{v_{r}} = (q_{rand} - q_{near})/\|q_{rand} - q_{near}\|$. To determine the potential biased direction $v_{extend}$ in which to extend $q_{near}$, we take a weighted sum of $\bm{v_{r}}$ and $\bm{v_{p}}$ as shown in Equation \ref{eq:weighted}.
\begin{equation}\label{eq:weighted}
    \hat{v}_{extend} = \lambda \bm{\hat{v}_r} + (1 - \lambda) \bm{\hat{v}_p}
\end{equation}
\begin{equation}\label{eq:lambda}
    \lambda = \frac{1}{\beta \frac{\max (0, F_{total})}{F_{att_{max}}}+1}
\end{equation}

Finally, we extend the nearest node in direction $\hat{v}_{extend}$ by a fixed distance $\delta$ to obtain the new node $q_{new} = q_{near} + \delta\hat{v}_{extend}$.

The parameter $\lambda$ helps to balance between following the direction of the random sample (exploration) and the direction of APF force (exploitation). We calculate $\lambda$ as inversely proportional to the ratio of the total force $F_{total}$ at nearest node (which is equal to $\vec{F}_{att} + \vec{F}_{rep}$) and the maximum attractive force $F_{att_{max}}$ (for scaling). This ensures that when the node is close to an obstacle or inside a local minima, $F_{total}$ is $\leq 0$, therefore $\lambda = 1$. So we completely follow the random direction $\bm{v_r}$ when in a local minima. This is specific to the lime picking scenario where the obstacles (leaves) are close to the goal and hence we want to explore when near the obstacles. 
In other scenarios i.e. $F_{total} > 0$, the denominator is $> 1$ and therefore we include a component of the APF to guide the extension of the tree.

The hyperparameter $\beta > 0$ can be tuned to adjust the weight for the APF direction  (Equation \ref{eq:lambda}). The higher the value of $\beta$ the larger the weight for extending the tree in the direction of the potential field gradient $\bm{v_p}$.

As we adjust $\lambda$ based on local forces (due to obstacles) we are able to adaptively search the C-space in a lime picking task. We bias towards the goal when away from the plant and explore when near the leaves. This allows us to find lower-cost solution in a limited number of iterations. 


\section{2-D Experiments}\label{sec:2d-exp}

\begin{table*}[t!]
\centering
\begin{tabular}{l|llll}
\hline
    \toprule
Algorithm  & 1000 iterations   & 2500 iterations & 5000 iterations \\
    \midrule
RRT$^*$& 971.19 (8.53) & 921.31 (9.88) & 846.09 (10.99)\\
APF-RRT$^*$ ($\beta = 1$)  & 880.43 (9.55) & 810.41 (10.54) & 726.93 (11.10)  \\
APF-RRT$^*$ ($\beta = 1.5$)  & 892.42 (8.47) & 835.43 (10.21) & 775.16 (11.53)  \\
APF-RRT$^*$ ($\beta = 2$)  & 900.51 (8.07) & 846.32 (8.89) & 768.36 (11.64)  \\
P-RRT$^*$ & 950.26 (8.74) & 874.25 (12.20) & 799.83 (13.64)  \\

  \bottomrule
\end{tabular}

\caption{Mean costs for paths found for our cost-based RRT$^*$ without potential bias (top row), with potential bias (second, third, fourth rows), and for our $Baseline$ P-RRT$^*$ (bottom row) for different numbers of iterations in the 2D environment. The numbers in parentheses indicate standard errors.}
\label{tab:results1}
\vspace{-1 ex}
\end{table*}

In this section, we visualize the behaviour of our proposed method in a 2-D environment for different values of the cost of collision with permeable obstacles and potential bias.

\begin{figure*}[t!]
    \begin{minipage}{.75\textwidth}
        \includegraphics[width=0.32\textwidth]{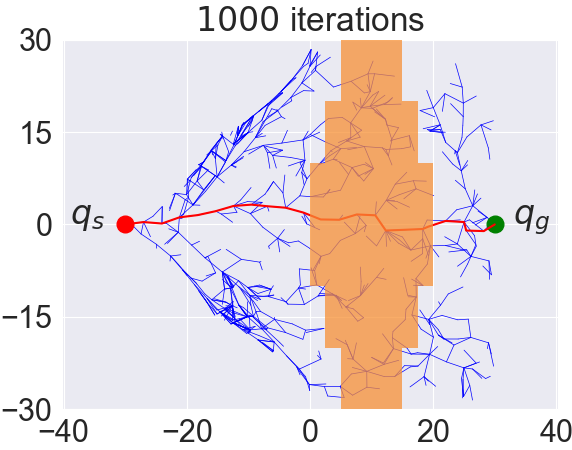}\label{fig:iters1000}
        \includegraphics[width=0.32\textwidth]{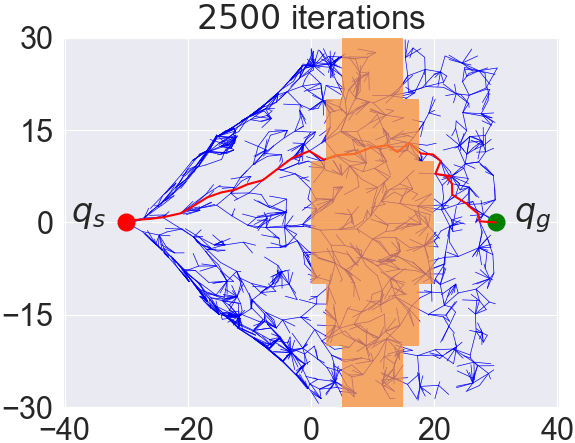}\label{fig:iters2500}
        \includegraphics[width=0.32\textwidth]{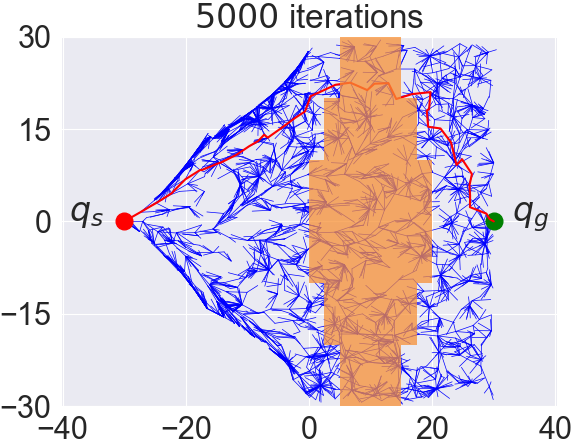}\label{fig:iters5000}
        \caption{Example paths found by our proposed method at different iterations for $C_{perm} = 100, \beta = 1.5$.}
        \label{fig:cost_iters}
    \end{minipage}%
    \begin{minipage}{0.25\textwidth}
        \centering
        \includegraphics[width=0.98\textwidth]{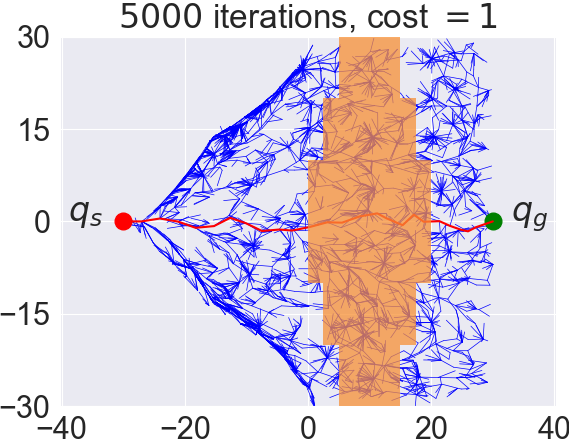}\label{fig:cost1}
        \caption{Example path for $C_{perm} = 1$}
        \label{fig:cost1}
    \end{minipage}
    \vspace{-1 ex}
\end{figure*}

\textbf{Wall environment.} 
Consider the environment shown in Figure \ref{fig:cost_iters} where the start ($q_s$) and goal ($q_g$) configurations are separated by a wall of permeable obstacles. We choose this environment to emulate the scenario where the path to a lime is completely blocked by leaves. Similar to how the density of foliage varies, the wall has wide and narrow portions. 
The shortest path between the start and goal, i.e., a straight line, would go through the widest region of the wall. Thus, it would have the largest number of nodes on the permeable obstacle and incur a high collision cost. 
We want our proposed method to find a path to the goal through either the top or bottom narrow portions of the wall where we would incur the least collision cost.

\textbf{Metric.} We compare the average cost of the path to goal $C_{path}(q_{goal})$ at $1000$, $2500$ and $5000$ iterations. The cost of a path $\tau$ is equal to the length of the path plus the incurred collision cost. In our proposed method the total collision cost incurred for path $\tau$ is dependent on the value of $C_{perm}$ and the number of nodes $n_{collision}$ on the path that are in collision with the permeable obstacle. Therefore, $C_{path}(q_{goal}) = d(\tau) + (C_{perm} \cdot n_{collision})$. 

We acknowledge that the metric depends on $n_{collision}$ for a given path $\tau$. While this dependency can make the metric inconsistent, our experiments empirically show that the cost of the generated paths decreases for increasing number of iterations. We select this metric for simplicity and we leave assigning collision costs to the edges at a fixed resolution for future work.



In the experiments, we set collision cost to a high value $C_{perm} = 100$, since we want to prioritize minimizing collisions. For consistency, we consider the same step size $\delta$ for all experiments and average the costs over $100$ trials.

\begin{figure}[t!]
    \centering
    \includegraphics[width=0.49\linewidth]{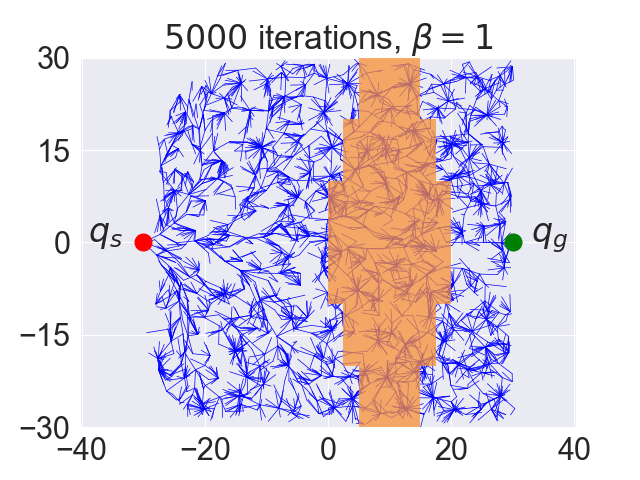}\label{fig:beta_1}
    \includegraphics[width=0.49\linewidth]{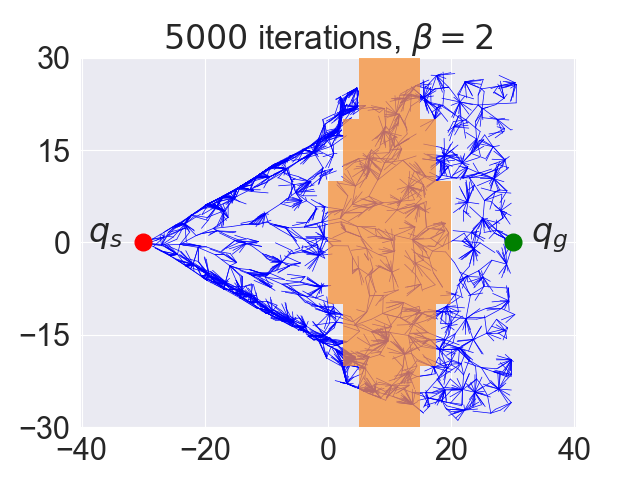}\label{fig:beta_2}
    \caption{Increasing $\beta$ increases the effect of potential bias for $K_{att} = 50$, $K_{rep} = 500$, $d_{obs} = 5.0$, $\delta = 3$ and $C_{perm} = 100$.}
    \label{fig:potential_bias}
    \vspace{-2 ex}
\end{figure}

\subsection{Effect of Potential Bias}\label{sec:apf-exp}

We now show that our proposed method of biasing the growth of the tree using an APF helps to find lower-cost paths in fewer iterations. Table~\ref{tab:results1} shows the costs of the paths found by our cost-based RRT* after $1000$, $2500$ and $5000$ iterations, averaged over $100$ trials, and the costs of the paths when we apply the potential bias with $K_{att} = 50$, $K_{rep} = 500$ and $\beta = 1$. We perform a two-way mixed ANOVA with the number of iterations as within-subjects factors and potential bias as the between-subjects factor. There was no statistically significant interaction between the number of iterations and the application of potential bias. The main effect of number of iterations showed a statistically significant difference in path cost at the different iterations, $(F(2, 396) = 166.985, p < 0.001)$. Post-hoc comparisons showed that the cost decreased for increasing number of iterations. The main effect of group also showed a statistically significant difference between the two algorithms $(F(1,198) = 88.103, p < 0.001)$. Therefore, we can see that we are able to find lower-cost paths by biasing the tree using our APF-RRT* approach.

\textbf{Parameter tuning.} Although our proposed method adapts the potential bias ($1 - \lambda$) based on the force due to the potential field (see Equations \ref{eq:weighted} \& \ref{eq:lambda}), the strength of the bias is dependent on the hyperparameter $\beta$ that needs to be tuned. Figure \ref{fig:potential_bias} shows the behaviour of our algorithm for different values of $\beta$. For $\beta = 1$ the random direction has a higher weightage ($\lambda$), and we can see that the tree explores most of the search space only missing out on the edges near the start state. As we increase $\beta$, the weight for random direction reduces and the growth of the tree is more biased towards the goal. However, this may or may not lead to improvements in the path cost depending on the obstacles in the environment. For $\beta = 1.5, 2$ the cost of paths after $1000$, $2500$ and $5000$ iterations are shown in Table~\ref{tab:results1} (third and fourth row), averaged over 100 trials. These path costs are larger than for $\beta = 1$, since increasing the potential bias too much biases the tree away from the narrow parts of the wall. Yet since the potential bias is adaptive, the path costs are still smaller than those without any potential bias.

\textbf{Setting the cost of permeable obstacles.} Fig.~\ref{fig:cost_iters} shows example paths found for $\beta=1.5$ and $C_{perm}=100$. We observe that as the number of iterations increases, APF-RRT* finds paths through the narrow parts of the wall. 

On the other hand, if the cost of collision with the wall is reduced to $1$ (see Figure \ref{fig:cost1}), the path with the least total cost after $5000$ iterations is the shortest length path that goes through the widest region of the wall. This is because the cost due to the length of the path dominates the cost incurred due to collision. Therefore, to minimize collision cost, the cost of collision with permeable obstacles ($C_{perm}$) must be set higher than the path lengths. For any significantly high value of collision cost, our proposed method will prioritize finding a path that incurs the least collision cost, with minimizing the path length as a secondary objective.

\subsection{Comparison of Potential-Biased RRT* Approaches}\label{sec:prrt-exp}


Lastly, we compare the performance of our APF approach to a state-of-the-art potential guided RRT* approach, P-RRT* \cite{qureshi2017prrt*}, that uses APF to bias the random sample. We wish to show that our proposed method is more effective in finding low-cost paths than P-RRT*. Therefore, we consider the approach of using APF to bias the random node as our $Baseline$ and compare the cost of paths after $5000$ iterations. 

\textbf{Implementation.} We use the same cost-based approach and potential field for both methods. The potential $U_{tot}$ at any configuration is calculated as in Section \ref{sec:apf}.



1) $Baseline$: We shift the random node along the APF gradient by a distance $\Delta$ for $k$ steps. At each step, the gradient is re-evaluated at the current configuration of the random node. We tune $\Delta$ and $k$ to achieve good performance. 2) Proposed APF-RRT*: We apply the potential bias to the nearest node with $\beta = 1$ as in Section \ref{sec:apf-exp}. 


\textbf{Comparison.} For $\Delta = 0.5$ and $k = 10$, the cost of paths found by  $Baseline$ (P-RRT*) after $1000$, $2500$ and $5000$ iterations, averaged over $100$ trials, are shown in Table~\ref{tab:results1} (bottom row). A two-tailed unpaired t-test showed a significant difference ($t(198) = 4.12$, $p<0.001$) between the cost of paths for $Baseline$ and our proposed method after $5000$ iterations. Thus, our proposed method can effectively bias the tree to achieve lower cost paths in the same number of iterations. Table~\ref{tab:results1} summarizes the performance of the tested algorithms.

\section{Robotic Lime Picking Experiments}\label{sec:robot}

In the previous section, we evaluated our proposed method in a simple environment. We now show how our proposed method can be used in a robotic lime picking task. 

\begin{figure}[hbt!]
    \centering
    \subfigure[Start configuration]{
    \includegraphics[width=0.29\linewidth]{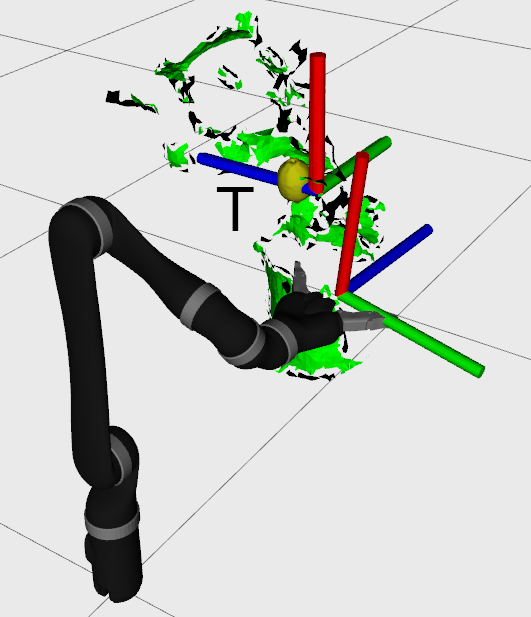}\label{fig:tsr}}
    \subfigure[$Baseline$]{
    \includegraphics[width=0.31\linewidth]{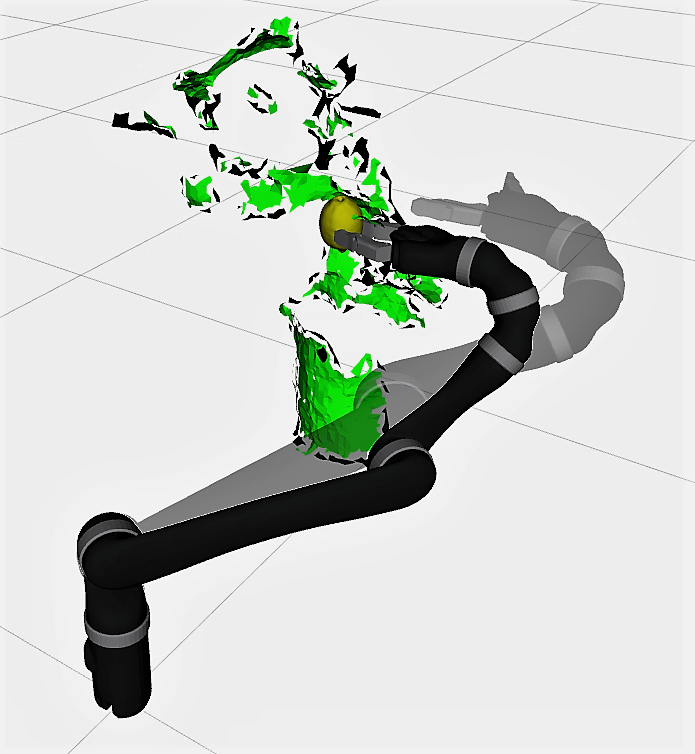}\label{fig:bad}}
    \subfigure[APF-RRT*]{
    \includegraphics[width=0.31\linewidth]{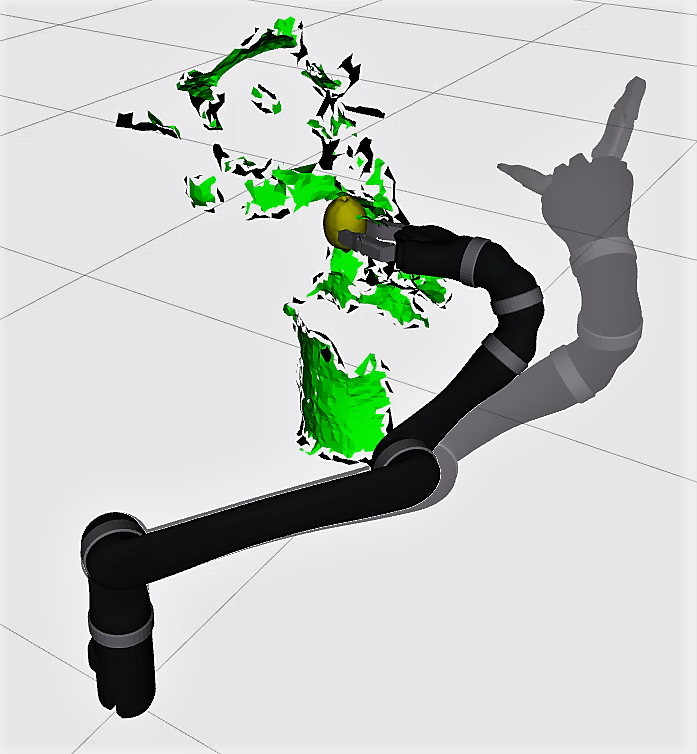}\label{fig:good}}
    \vspace{-1.5ex}
    \caption{The plant point cloud is shown in green, the lime is yellow, and the robot in dark grey. (a) We select a grasp pose $T$ that approaches the lime through the leaves. (b) Path found by $Baseline$ incurs higher collision cost as it approaches the lime from the front which is blocked by leaves. (c) APF-RRT* explores near the leaves to find an alternate path from the side that has less collision with the leaves.}
    \label{fig:simulation}
    \vspace{-1 ex}
\end{figure}

\textbf{Implementation.} The environment is set up with a 6-DOF Kinova Gen2 robot arm in front of an artificial lime plant (see Figure \ref{fig:simulation}). We load the plant as a point cloud where we manually label the penetrable (foliage) and impenetrable (stem) obstacles. To measure collision with the point cloud, we use the FCL (Flexible Collision Library) \cite{pan2012fcl}.
Since for high-dimensional problems it may not be possible to map obstacles to the configuration space, we approximate the computation of repulsive potentials by taking the shortest workspace distance between the robot and the point cloud.

In this experiment, we load the lime as a 3D object and hand-pick an end-effector pose $T$ - that forces the robot to move through the leaves for grasping the lime. 

\textbf{Evaluation.} We first evaluate whether our proposed APF approach can find lower cost paths to $T$ compared to using no potential bias. We measure the path costs at $1500$ iterations for $C_{perm} = 100$, $\delta = 0.1$ and $\beta = 1.0$ and averaged over $100$ trials. An unpaired two-tailed t-test showed a significant decrease ($t(182.25) = -6.39$, $p < 0.001$) in the cost of paths with the proposed method. Therefore, we see that our proposed method helps to find lower cost paths to limes.

We also compare our proposed method to $Baseline$ (P-RRT*) as in Section \ref{sec:prrt-exp}. We tune the $Baseline$ parameters to $\Delta = 0.1$ and $k = 2$. We again measure the cost of path to $T$ at $1500$ iterations for $C_{perm} = 100$, $\delta = 0.1$ and $\beta = 1.0$, and average the cost over $100$ trials. An unpaired two-tailed t-test showed a significant decrease ($t(186.07) = 7.63$, $p < 0.001$) in the cost of paths with the proposed method, for the same $K_{att}$ and $K_{rep}$. Table~\ref{tab:results2} summarizes these results.

Thus, we see that our proposed method is more effective at finding lower-cost paths than the $Baseline$. This is because our approach of reducing the potential bias when the robot is near the leaves allows us to explore the plant foliage and find better paths.




\begin{table}[t!]
\centering
    \begin{tabular}{l|l}
\hline
    \toprule
Algorithm  &  1500 iterations\\
    \midrule
RRT$^*$& 822.75 (50.06) \\
APF-RRT$^*$   & 422.62 (36.98)\\
P-RRT$^*$ & 886.77 (47.90)  \\

  \bottomrule
\end{tabular}
\vspace{1ex}
\caption{Mean costs of paths found by our cost-based RRT$^*$ without potential bias (top row), with potential bias (second row), and by P-RRT$^*$ (bottom row) for 1500 iterations in the lime picking simulation. The numbers in parentheses indicate standard errors.}
\label{tab:results2}
\vspace{-3ex}
\end{table}

\textbf{Real-world demonstration.}
In addition to the simulation experiments, we also demonstrate the utility of our approach in a real-world robotic lime picking task (refer to the supplementary video).
\section{Conclusion}\label{sec:con}
Our work focuses on the problem of accounting for collision with the leaves while picking limes. We consider the leaves as permeable obstacles, and adapt the RRT* algorithm by incorporating the collision cost while selecting the best parent for a new node and rewiring its neighbours. We also propose using an APF that effectively biases the growth of the tree to explore areas near the leaves. Our experiments show that our method can help to reduce collision with the leaves during lime picking, and can motivate further research in cost-based planning for robotic fruit harvesting.

A limitation of our method is that it requires setting the value of $C_{perm}$ to balance the trade-off between minimizing the path length and the cost of collision with permeable obstacles. Moreover, as it is difficult to quantify the damage caused to the plant foliage \cite{bac2014harvesting}, we assume the damage to be proportional to the incurred collision cost. Future work can look into automatically learning the collision costs for different penetrable obstacles from demonstration, as well as extending the notion of permeable obstacles to other domains, such as tabletop arrangement tasks~\cite{batra2020rearrangement}. 
 




\bibliographystyle{IEEEtran}
\bibliography{references}

\end{document}